\pdfoutput=1

\documentclass[11pt]{article}

\usepackage[preprint]{acl}

\usepackage{times}
\usepackage{latexsym}

\usepackage[T1]{fontenc}

\usepackage[utf8]{inputenc}

\usepackage{microtype}

\usepackage{inconsolata}

\usepackage{graphicx}
\usepackage{xspace,mfirstuc,tabulary}
\usepackage{booktabs}
\usepackage{multirow}
\usepackage{stfloats}
\usepackage{dsfont}
\usepackage{tabularx}
\usepackage{subcaption}
\usepackage{amssymb}
\usepackage{amsmath}
\usepackage{hhline} 
\usepackage{ulem}
\usepackage{caption}
\title{E2TP: Element to Tuple Prompting Improves Aspect Sentiment Tuple Prediction}

 \author{Mohammad Ghiasvand Mohammadkhani\thanks{\footnotesize{Work was done while Mohammad was a third-semester undergraduate student.}} \and Niloofar Ranjbar \and  Saeedeh Momtazi \\
         Amirkabir University of Technology, Iran \\ \texttt{\{mohammad.ghiasvand, nranjbar, momtazi\}@aut.ac.ir} \\ 
           }

\begin{document}
\maketitle
\begin{abstract}
Generative approaches have significantly influenced Aspect-Based Sentiment Analysis (ABSA), garnering considerable attention. However, existing studies often predict target text components monolithically, neglecting the benefits of utilizing single elements for tuple prediction. In this paper, we introduce \textit{\textbf{E}lement \textbf{to} \textbf{T}uple \textbf{P}rompting} (E2TP), employing a two-step architecture. The former step focuses on predicting single elements, while the latter step completes the process by mapping these predicted elements to their corresponding tuples. E2TP is inspired by human problem-solving, breaking down tasks into manageable parts, using the first step’s output as a guide in the second step. Within this strategy, three types of paradigms, namely E2TP(\(diet\)), E2TP(\(f_1\)), and E2TP(\(f_2\)), are designed to facilitate the training process. Beyond dataset-specific experiments, our paper addresses cross-domain scenarios, demonstrating the effectiveness and generalizability of the approach. By conducting a comprehensive analysis on various benchmarks, we show that E2TP achieves new state-of-the-art results in nearly all cases.\footnote{\footnotesize{Code and data released at  \url{https://github.com/mghiasvandm/E2TP}}}
\end{abstract}

\section{Introduction}

In recent years, ABSA has garnered significant attention as a nuanced task within the broader field of sentiment analysis. ABSA is designed to forecast tuples of sentiment elements embedded within a specified input. The core components are categorized into one of the following types: aspect term ($a$), aspect category ($c$), opinion term ($o$), and sentiment polarity ($s$) \citep{zhang2022survey}.
Initially, ABSA research primarily focused on identifying single elements, such as aspect terms \citep{liu2015fine,ma2019exploring}, aspect categories \citep{zhou2015representation}, and sentiment polarities \citep{chen2017recurrent}. More recent studies, however, have shifted focus to extracting triplets and quadruplets, as evidenced by works on Aspect Sentiment Triplet Extraction (ASTE) \citep{peng2020knowing}, Target Aspect Sentiment Detection (TASD) \citep{wan2020target}, Aspect Sentiment Quad Prediction (ASQP)\citep{zhang2021aspect}, and Aspect Category Opinion Sentiment (ACOS) \citep{cai2020aspect}, each targeting a distinct ABSA task. The formats of these tasks for the input sentence ``The sushi is tasty'' are illustrated in Table \ref{tab:tformat}. 

As \citet{zhang2022survey} noted, ABSA methodologies typically fall into distinct categories, including sequence-to-sequence modeling (seq2seq), machine reading comprehension, and sequence labeling. However, although all ABSA tasks can be reformulated as seq2seq problems, various studies discuss designing sequences comprising different sentiment elements. For instance, \citet{yan2021unified} introduced the use of class index, while \citet{zhang2021aspect2} and \citet{liu2021solving} employed natural language techniques. \citet{gao-etal-2022-lego} concentrated on formulating prompt instructions for pair extractions using T5 \citep{t5cite} sentinel tokens, and MvP by \citet{gou-etal-2023-mvp} explored multi-view prompting for sequencing input and target elements in different orders. Recent interest has surged in exploring cross-domain ABSA settings alongside traditional in-domain configurations due to practical challenges in labeling and preparing diverse data. There's a crucial demand for models effective across various domains. While many approaches excel in supervised in-domain settings, their performance in cross-domain scenarios needs validation, particularly in advanced ABSA tasks like ASTE. Bridging this gap, \citet{deng2023bidirectional} proposed BGCA, a novel data augmentation strategy leveraging noisy generated data to enhance target domain knowledge in source data. BGCA has achieved state-of-the-art performance, showcasing potential to overcome challenges and meet the demand for versatile ABSA models.

\begin{table*}[t]
	\captionsetup{font=footnotesize}
	\centering
	\begin{tabular}{cccl}
		\toprule
		Task  & Output \\
		\midrule
		Aspect Sentiment Triplet Extraction (ASTE)  & sushi, tasty, positive $(a, o, s)$\\
		Target Aspect Sentiment Detection (TASD) &  sushi, food quality, positive $(a, c, s)$ \\
		Aspect Sentiment Quad Extraction (ASQP)  & sushi, food quality, tasty, positive $(a, c, o, s)$ \\
		Aspect Category Opinion Sentiment (ACOS)  & sushi, food quality, tasty, positive  $(a, c, o, s)$ \\
		\bottomrule
	\end{tabular}
	\caption{ABSA Task Target Formats. Notably, the target formats for ASQP and ACOS tasks are the same, but ACOS pays more attention to the implicit role of aspect terms and opinion terms than ASQP does.}
	\label{tab:tformat}
\end{table*}

Previous research has predominantly focused on generating the target sequence in a single step, overlooking the potential benefits of employing a two-step approach. This method utilizes the guidance of single elements for tuple prediction from different paths and starting points. Inspired by human problem-solving intuition, this strategy involves breaking down the problem into smaller, more manageable subtasks.

To tackle the mentioned challenge and leverage our observations, we introduce E2TP, which utilizes a two-step architecture. Initially, our approach predicts single elements independently as potential candidates for gold elements. Subsequently, in the second step, it maps and finalizes tuples associated with the selected elements. Within the E2TP framework, we've customized the modeling of ABSA tasks to achieve two primary goals: i) predicting the maximum number of correct single elements, and ii) accurately mapping and completing tuple predictions based on the predicted elements.
The E2TP framework offers several advantages by breaking down problems into simpler tasks, training separate expert models for each step, facilitating easier enhancement, and filtering out incorrect candidates. It generates various paths with different fixed starting points for constructing target tuples and selects the most probable ones through aggregation, providing valuable flexibility in predicting specific element types from complex inputs.

To facilitate the implementation of the E2TP strategy, three innovative and potent paradigms have been developed: E2TP(\(diet\)), which elucidates the main idea of the paper and is notably data-efficient; E2TP(\(f_1\)) and E2TP(\(f_2\)), which boast higher data augmentation rates. Besides the variance in data augmentation strategies among the paradigms, they also employ different prompting templates. In summary, our paper makes several key contributions:

\begin{enumerate}
	\item We introduce E2TP, a simple yet effective two-step prompting framework that integrates three distinct paradigms to streamline ABSA processes, thereby improving predictions. Additionally, E2TP's intuition can be applied to other Natural Language Processing (NLP) tasks involving representing outputs as tuples of elements.
	\item By integrating the BGCA data augmentation strategy into E2TP, we enhance performance in the ASTE task under cross-domain configuration, effectively addressing a recent and challenging issue in the field.
	\item Comprehensive experimental results demonstrate that E2TP significantly advances the state-of-the-art in nearly all cases.
\end{enumerate}
\section{Related Works}
\label{sec:relworks}
\subsection{Aspect-based Sentiment Analysis}
\label{absa}
The ABSA task garners NLP interest due to its complexity. Early research focused on single term or pair prediction such as the proposed models by \citet{liu2015fine,ma2019exploring} which concentrated on aspect term extraction, and also the proposed models by \citet{huang2018aspect,luo2019doer,chen2020aspect} which focused on pair extraction tasks. However, with the growth of prompt studies and the advent of new transformers such as T5, BERT \citep{devlin-etal-2019-bert}, and BART \citep{lewis-etal-2020-bart}, the focus has shifted towards triplet or quadruplet predictions which are known as ASTE, TASD, ASQP and ACOS.

\subsection{Generative ABSA}
\label{gabsa}
Generative ABSA is gaining attention, particularly with the emergence of new transformer models and prompt-based studies. \citet{zhang2021towards2} introduce GAS, treating ABSA as a seq2seq challenge with two paradigms. Another study by \citet{zhang2021aspect2} presents Paraphrase, which rephrases sentiment tuples and applies label semantics. UIE, as presented by \citet{lu2022unified}, introduces a unified pre-training framework trained on a wide range of ASTE instances. \citet{gao-etal-2022-lego} develop LEGO-ABSA, utilizing T5 sentinel tokens and solving advanced tasks by element pairs extraction. \citet{mao2022seq2path} introduce Seq2Path, predicting sentiment tuples through tree paths. \citet{hu2022improving} propose DLO, considering the order of elements in target text and evaluating multiple templates for quadruplet tasks. MvP extends DLO by sequencing inputs and training on augmented samples to tackle the one-to-many gap. \citet{deng2023bidirectional} present BGCA, a data augmentation strategy for target domain knowledge distillation, and CONTRASTE proposed by \citet{mukherjee-etal-2023-contraste}, utilizing contrastive pre-training based on aspect-based prompts. Nevertheless, The potential of using single elements for predicting tuples from various fixed starting points remains yet to be fully explored.

\subsection{Other ABSA Approaches}
\label{relw3}
In this case, as the main focus of this work is on generative and specifically seq2seq approaches, we only mention very few other methods from recent studies that follow different approaches for advanced ABSA tasks. Extract-Classify, proposed by \citet{cai2021aspect}, introduces a discriminative approach for the ACOS task, involving extracting aspect-opinion pairs and then classifying them. \citeauthor{xu2021learning} proposed Span-ASTE, which uses spanning techniques to predict tuples  span-level interactions.  \citeauthor{liang2023stage} presented STAGE with a novel span tagging method recognizing diverse span roles and a greedy inference scheme. BMRC by \citet{chen2021bidirectional} and RoBMRC by \citet{liu2022robustly} use machine reading comprehension for prediction, with RoBMRC adding features like span matching and exclusive classifiers for enhancement. \citet{yuan2023encoding} suggests a syntax-aware transformer for triplet extraction, integrating dependency type knowledge into graph neural networks. TAGS by \citet{xianlong2023tagging} combining sequence labeling with a generative model for improved predictions and semantic alignment.

\subsection{Cross-domain ABSA}
Advancements in in-domain states highlight the significance of cross-domain ABSA, essential for models to effectively generalize across various domains, especially for complex tasks like ASTE. Although several approaches exist for cross-domain ABSA, more advanced tasks such as ASTE in cross-domain settings have not been extensively explored until recently. BGCA's notable contribution lies in its thorough examination, evaluating previous models like GAS, RoBMRC, and Span-ASTE, and introducing a novel bidirectional generative framework that doesn't rely on task-specific designs or external data.

\section{Methodology}

\begin{figure*}[t]
	\captionsetup{font=footnotesize}
	\centering
	\includegraphics[width=\textwidth]{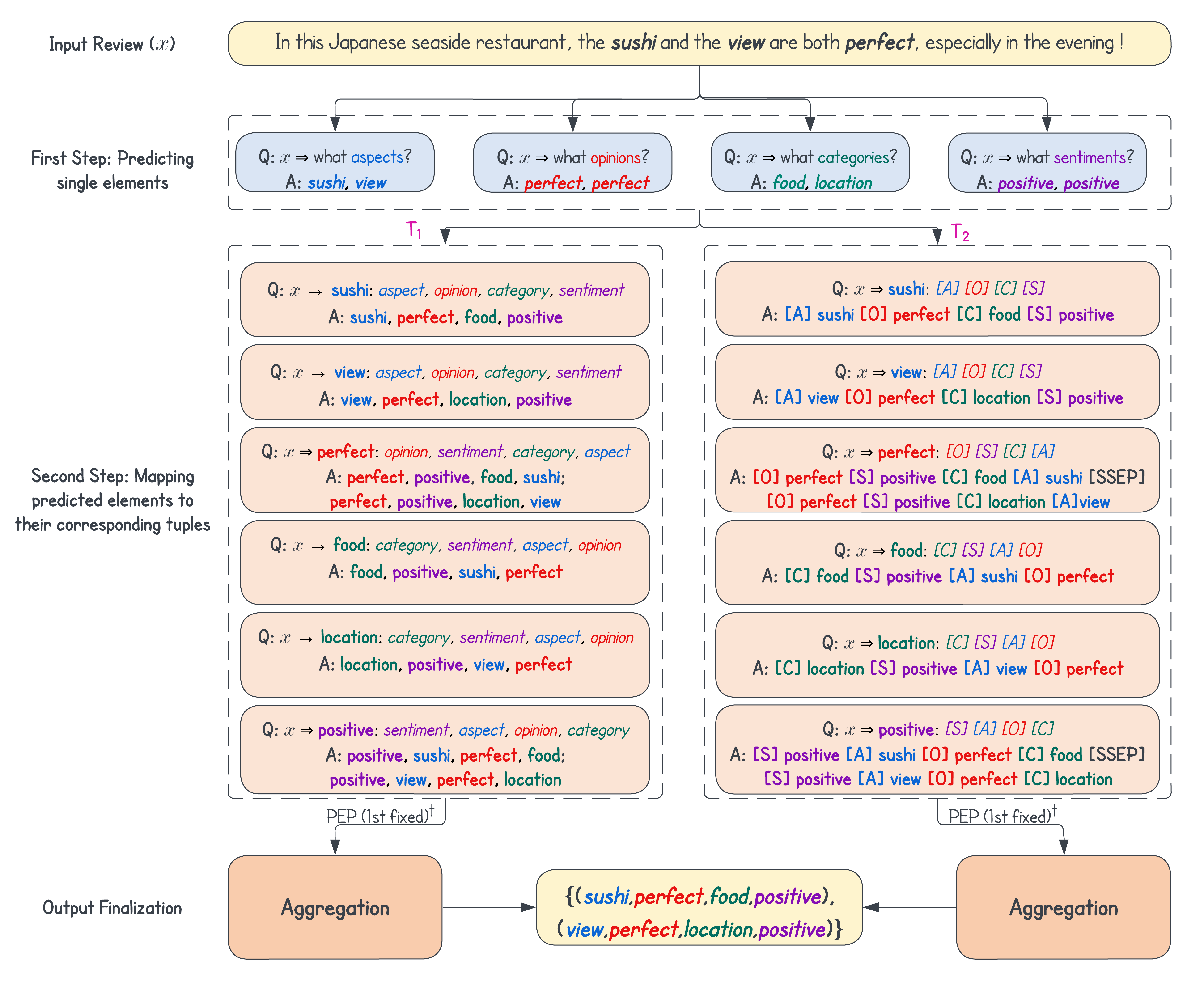}
	\caption{E2TP Framework Illustration. † indicates prompt elements permutation (1st fixed) described in section \ref{sec:pepf}}
	\label{fig:fig1}
\end{figure*}

\label{sec:method}
This section provides a detailed overview of the operational procedure of the E2TP model. It discusses the three paradigms and explains the framework outlined in Figure \ref{fig:fig1}, which serves as an illustrative example for the quadruplet task.

\subsection{Problem Definition}
In this part, we represent the default quadruple task approach, which can seamlessly extend to triplet tasks with slight adjustments. Our formal definition of the task is as follows: when provided with an input sentence denoted as $x$, the goal is to forecast all sentiment tuples $T = \{(a_i, c_i, o_i, s_i)\}_{i=1}^{|T|}$ in $x$, where each tuple encompasses an aspect term ($a_i$), aspect category ($c_i$), opinion term ($o_i$), and sentiment polarity ($s_i$). Similar to prior research \citep{zhang2021aspect2,gou-etal-2023-mvp}, we replace the “NULL” label of aspect term $a_i$ with “it”. However, contrary to previous studies, we maintain “positive”, “neutral”, and “negative” as sentiment polarity labels and do not apply label semantics according to conflicts between the labels “great”, “bad”, and “ok” and the words within the sentence.

\subsection{Training}

E2TP uses a two-step prompting mechanism for accurate sentiment tuple predictions. It involves designing distinct targets and inputs for each step and initializing parameter $\theta$. Each step employs a T5 model, fine-tuned independently  as the backbone model to minimize the Cross-Entropy loss function during training:
\begin{equation}
	\mathcal{L}(x,y) = -\sum_{t=1}^{n} \log p_{\theta}(y_{t} | x, y_{<t})
\end{equation}
where $n$ represents the length of the target sequence $y$, and $y_{<t}$ denotes the tokens generated previously.

\subsubsection{First Step Input and Target Format} 
As mentioned previously, the process involves two steps; The first step's architecture remains constant, while variations occur in the second step across paradigms. At the outset, the objective is to forecast elements such as aspects, categories, opinions, or sentiments, depending on the specifics of the task. If \(n\) represents the count of distinct element types within the task, a new set of data points \((x_i, y_i)_{i=1}^n\) is derived from \(x\). With \(Q\) and \(A\) defined as \(Q = \{ t_1: \text{``aspects''}, t_2: \text{``categories''}, t_3: \text{``opinions''}, t_4: \text{``sentiments''} \}\) and \(A = \{ t_1: \text{``sushi, view''}, t_2: \text{``food, location''}, t_3: \text{``perfect, perfect''}, t_4: \text{``positive, positive''} \}\), the structuring of input and target for each \(i\) from 1 to \(n\) is as follows:
\begin{gather*}
	x_i\!: x \Rightarrow \text{what } Q[\text{t}_i]?\\
	\vspace{-0.5mm}
	y_i\!: A[\text{t}_i]
\end{gather*}

The primitive dataset undergoes annotation for each data point, resulting in $D_1$ as the final dataset. Subsequently, the model undergoes fine-tuning on $D_1$ for a seq2seq task, extending the initial model's progress.

\subsubsection{Second Step Input and Target Format}

In the second step, three paradigms are formulated. During this phase of training, similar to the first step, the primitive dataset consists of pairs $\left\{(x_i, y_i)\right\}_{i=1}^{N}$, where $N$ represents the total number of data points. Here, $x_i$ and $y_i$ represent the input and target texts respectively. It is important to note that this step operates independently from the first step, allowing for concurrent training of the two models, thereby avoiding an increase in time complexity. In this step, for each input labeled as \(x_i\), we can assign \(y_i\) to a set $T = \left\{(a_i, c_i, o_i, s_i)\right\}_{i=1}^{|T|}$. For each $x_i$, we construct $E = \left\{a_i\right\}_{i=1}^{|T|} \cup \left\{c_i\right\}_{i=1}^{|T|} \cup \left\{o_i\right\}_{i=1}^{|T|} \cup \left\{s_i\right\}_{i=1}^{|T|}$, forming a set of unique elements. During inference, single elements are accessed instead of tuples, enabling the direct formation of set \(E\) from the first step predictions. Initially, we outline our designed prompting templates and introduce prompt elements permutation (1st fixed), followed by a comparison of the three paradigms.

\textbf{First Template (T\textsubscript{1}):} This template begins by preparing prompt elements for each \(e \in E\). In this quadruplet prediction task, \(t_p\) is defined as the task prompt, representing the sequence of different element types corresponding to each \(e \in E\). The type of each element, denoted as \(t_e\), guides the design process. The set \(L\) is initialized with existing element labels within the task: \(L = \{aspect, category, opinion, sentiment\}\). Elements of \(L\) are concatenated and separated by ``, ''. If \(t_e\) is not the first item, its position is swapped with the first item to ensure precedence. Consequently, \(t_p\) is defined accordingly. The input \(x_e\) for the new dataset is then determined.

\[
x_e:= x_i \rightarrow e\!\!: t_p
\]

To ensure the consistent generation of \(e\) as the initial element of all prediction tuples, \(e\) precedes \(t_p\), with \(t_e\) becoming the first word of \(t_p\). Additionally, the symbol “\(\rightarrow\)” is changed to “\(\Rightarrow\)” when \(e\) appears more than once within \(T\), aiding in symbol differentiation during inference. This change prompts the model to produce a single tuple when encountering “\(\rightarrow\)”, while “\(\Rightarrow\)” triggers the generation of multiple tuples, reflecting the multiplicity indication.


\textbf{Second Template (T\textsubscript{2}):}
This template differs from the previous one in two main ways. Firstly, it exclusively uses “\(\Rightarrow\)” symbol instead of “\(\rightarrow\)”. Secondly, it adopts markers to represent element types, as described in \citet{gou-etal-2023-mvp}: {\fontfamily{lmvtt}\selectfont [A]} for aspect term, {\fontfamily{lmvtt}\selectfont [C]} for aspect category, {\fontfamily{lmvtt}\selectfont [O]} for opinion term, and {\fontfamily{lmvtt}\selectfont [S]} for sentiment polarity. Each output element is prefixed by its corresponding marker, with different tuples separated by “ {\fontfamily{lmvtt}\selectfont [SSEP]} ”.

\textbf{Prompt Elements Permutation (1st fixed):}
\label{sec:pepf}
After constructing the new dataset, named \(D'\) using either the first or second template, for the second step, the E2TP approach for input reconstruction involves anchoring the first prompt element in the task prompt and permuting the remaining elements to generate new orders within the same task prompt. For instance, in the case of ASTE, if the task prompt begins with the word ``$aspect$'' or the marker ``{\fontfamily{lmvtt}\selectfont [A]}'' fixed at the start, two potential augmentation options could be ``$aspect, opinion, sentiment$'' and ``$aspect, sentiment, opinion$'' or ``{\fontfamily{lmvtt}\selectfont [A]} {\fontfamily{lmvtt}\selectfont [O]} {\fontfamily{lmvtt}\selectfont [S]}'' and ``{\fontfamily{lmvtt}\selectfont [A]} {\fontfamily{lmvtt}\selectfont [S]} {\fontfamily{lmvtt}\selectfont [O]}'' based on the prompting design mentioned in templates T\textsubscript{1} or  T\textsubscript{2}. By appending these sequences after the colon, two annotations for that data point can be generated. Consequently, despite the inference stage wherein only the input is accessible and permutations are done only within the input components, in the training process, the target elements are also permuted accordingly to reflect the order of task prompt elements. In a quadruplet task, after fixing the first item, each data point in  \(D'\) has \((4-1)! = 6\) possibilities, and in a triplet task, \((3-1)! = 2\) possibilities. In this work, two approaches have been applied to select data among the created possibilities based on the permutations. These approaches were used to construct $D_2$ as the final dataset for the second step, and then the T5 model was fine-tuned on $D_2$. The two approaches are as follows:

\begin{itemize}	
	\item \textbf{Diet:}
	In this scenario, a random seed is initialized, and one permutation is randomly selected from the 6 or 2 possibilities for the data points within \(D'\). Notably, data points in \(D'\) with the same first element in their task prompt elements must have the same permutation order selected.
	\item \textbf{Full Selection:} This approach considers all possible permutations for each data point inside \(D'\), effectively integrating E2TP with an adapted version of MvP intuition to achieve a higher data augmentation rate.
\end{itemize}

\subsubsection{\textbf{Paradigms:}}

\begin{table*}[t]
	\captionsetup{font=footnotesize}
	\footnotesize
	\centering
	\begin{tabularx}{\textwidth}{l||*{2}{>{\centering\arraybackslash}X}|*{4}{>{\centering\arraybackslash}X}|*{2}{>{\centering\arraybackslash}X}|*{2}{>{\centering\arraybackslash}X}}
		\toprule[1.0pt]
		\multirow{2}{*}{\textbf{Methods}}&\multicolumn{2}{c|}{\textbf{TASD}}& \multicolumn{4}{c|}{\textbf{ASTE}}&\multicolumn{2}{c|}{\textbf{ASQP}}& \multicolumn{2}{c}{\textbf{ACOS}}\\
		&\textbf{R15}&\textbf{R16}&\textbf{L14}&\textbf{R14}&\textbf{R15}&\textbf{R16}&\textbf{R15}& \textbf{R16}&\textbf{Lap}&\textbf{Rest}\\
		\midrule[0.5pt]
		BMRC \citep{chen2021bidirectional} & -& -&57.82& 67.99 & 60.02 & 65.75 &  - & - &  - & - \\
		Extract-Classify \citep{cai2021aspect}&-&  - & - & - & - & - & 36.42 & 43.77 & 35.80 & 44.61 \\
		GAS \citep{zhang2021towards2} &60.63& 68.31 &58.19 & 70.52 & 60.23 & 69.05 & 45.98 & 56.04 & - & - \\
		Span-ASTE \citep{xu2021learning}&-& - & 59.38 & 71.85 & 63.27 & 70.26 & - & - & - & - \\
		Paraphrase \citep{zhang2021aspect2} &63.06& 71.97 & 61.13 & 72.03 & 62.56 & 71.70 & 46.93 & 57.93 & 43.51 & 61.16 \\
		UIE  \citep{lu2022unified}&-& - & 62.94 & 72.55 & 64.41 & 72.86 & - & - & - & - \\
		DLO \citep{hu2022improving}&62.95& 71.79 &61.46 & 72.39 & 64.26 & 73.03 & 48.18 & 59.79 & 43.64 & 59.99 \\
		Seq2Path \citep{mao2022seq2path}&63.89& 69.23 & 62.22 & 73.75 & 64.02 & 73.40 & - & - & 42.97 & 58.41 \\
		LEGO-ABSA \citep{gao-etal-2022-lego}&62.30& 71.80 & 62.20 & 73.70 & 64.40 & 69.90 & 46.10 & 57.60 & - & - \\
		MvP \citep{gou-etal-2023-mvp}&64.53& 72.76 &63.33 & 74.05 & 65.89 & 73.48 & 51.04 & 60.39 & \smash{\underline{43.92}} &\smash{\underline{61.54}}\\
		TAGS \citep{xianlong2023tagging}  & - & - & 64.53 & 75.05 & 67.90 & \textbf{76.61} & - & - & - & - \\
		\midrule[0.5pt]
		E2TP(\(diet\)) &64.79& 72.66 & \textbf{67.31} & 74.51 & 67.47 & \smash{\underline{76.19}} & 51.70 & \textbf{62.90} &\textbf{45.00}& \textbf{61.89}  \\
		E2TP(\(f_1\)) &\textbf{65.80}& \textbf{73.17} & 65.61& \smash{\underline{75.28}}& \smash{\underline{68.08}}& 75.74 & \smash{\underline{51.94}}  & \smash{\underline{62.57}} & - & - \\
		E2TP(\(f_2\)) &\smash{\underline{65.39}}& \smash{\underline{72.95}}  & \smash{\underline{66.74}} &\textbf{75.53} &\textbf{68.77}& 75.46 & \textbf{52.11} &62.10&-&-\\
		\bottomrule[1.0pt]
	\end{tabularx}
	\caption{Results are presented for 10 datasets across TASD, ASTE, ASQP, and ACOS tasks. The best results are highlighted in bold, and the second best are underlined. All results from other papers were sourced from \citet{gou-etal-2023-mvp} or \citet{xianlong2023tagging}
	}
	\label{tab:table3}
\end{table*}

The section outlines three paradigms used in the study to achieve core objectives by leveraging task prompts and selection methods for data augmentation. These paradigms aim to explore strategies enhancing model effectiveness in tuple prediction.
\begin{itemize}	
	\item \textbf{First Paradigm (\(diet\)):}
	This paradigm represents the core idea of the paper; For the second step, it uses T\textsubscript{1} template with a slight modification, replacing the symbol “$\Rightarrow$” instead of “$\rightarrow$” and it also utilizes the diet method.
	
	\item \textbf{Second Paradigm (\(f_1\)):}
	In this paradigm, the full selection method is applied under the T\textsubscript{1} template and is named \(f_1\).
	
	\item \textbf{Third Paradigm (\(f_2\)):}
	This paradigm, utilizes the T\textsubscript{2} template and performs the full selection method and it is named \(f_2\).
\end{itemize}
Details on the original datasets and comparisons between MvP, E2TP(\(diet\)), and E2TP(\(f_1\)) or E2TP(\(f_2\)) regarding data augmentation rate are provided in the Appendix \ref{ap:third}.

\subsubsection{Cross-domain}

In cross-domain training, we begin by applying the BGCA data augmentation strategy outlined in the Appendix \ref{sec:bgca}. However, instead of training a new model with a mix of primitive and new data following the final step of the BGCA approach, we integrate the mixed data into the E2TP framework to implement E2TP in a cross-domain context.

\subsection{Inference}

During inference, unlike the training process, the two models perform dependently. The initial model predicts single elements and uses a lightweight constraint decoding to ensure correct token types. This forms a new dataset \(D_1'\) for further evaluation in the second step.

To continue evaluation, we utilize \(D_1'\) as our dataset and, based on different approaches in the second step, we create inputs for the second step. Notably, in the \(diet\) paradigm's inference step, task prompts are randomly selected from permutations with the same seed as in training. Outputs from the second step model are then used to create \(D_2'\). Following the input design from previous sections, each input in \(D_2'\) consists of multiple parts: the review sentence, represented by $x$, the single element, represented by $e$, and the task prompt, along with either “$\rightarrow$” or “$\Rightarrow$” with a colon. Upon examining \(D_2'\), we group the dataset by shared review sentences and establish an aggregation method for each group, which can be generalized to finalize the output for other groups. Within each group, we validate the generated tuples against the specified prompting template's syntactic format, discarding those that don't conform. Then, we count the inputs in the group as $k$ and initialize a set \(T_k'\) for each input, assigning the generated tuples to \(T_k'\). The final aggregation is performed using the equation below, and \(T'\) represents the finalized outputs for the specific review sentence within the group.

\begin{equation}
	T' = \left\lbrace t\in \bigcup_{i=1}^{k} T'_{i} \bigg| \sum_{i=1}^{k} \mathds{1}_{T'_{i}}(t) > m \right\rbrace 
\end{equation}

The equation includes a variable \(m\) that adjusts according to the task. Another hyperparameter, \(d\), isn't explicitly stated in the formula but serves to modify the process. If \(T'\) stays vacant after applying the equation, \(m\) decreases by one, and the process repeats, with a maximum of \(d\) iterations if \(T'\) remains empty.
\section{Experiments}
\label{sec:experiments}
Detailed experimental settings are available in Appendix \ref{sec:imp}.

\begin{table*}[t]
	\centering
	\captionsetup{font=footnotesize}
	\footnotesize
	\begin{tabular}{l||cccccc}
		\toprule[1.0pt] 
		\textbf{Methods}&\textbf{R14$\rightarrow$L14} & \textbf{R15$\rightarrow$L14} & \textbf{R16$\rightarrow$L14} &\textbf{ L14$\rightarrow$R14} &\textbf{ L14$\rightarrow$R15} & \textbf{L14$\rightarrow$R16} \\ \midrule[0.5pt]
		RoBMRC \citep{liu2022robustly} & 43.90               & 40.19               & 37.81               & 57.13               & 45.62               & 52.05               \\ 
		Span-ASTE \citep{xu2021learning}                    & 45.83               & 42.50               & 40.57               & 57.24               & 49.02               & 55.77               \\ 
		GAS  \citep{zhang2021towards2}                         & 49.57               & 43.78               & 45.24               & 64.40               & 56.26               & 63.14               \\ 
		BGCA  \citep{deng2023bidirectional}                     & 53.64               & 45.85               & 47.28       & 65.27               & 58.95               & 64.00               \\  \midrule[0.5pt]
		E2TP(\(diet\)) &  54.28 & 46.23 & \smash{\underline{47.39}} & \textbf{69.71} & 61.71 & 67.24  \\
		E2TP(\(f_1\))                      & \textbf{55.18}      & \smash{\underline{46.66}}          & \textbf{48.40}      & 68.42          & \smash{\underline{61.94}}          & \textbf{67.79}      \\ 
		E2TP(\(f_2\))                      & \uline{54.70}          & \textbf{46.74}      & 46.74               & \smash{\underline{69.24}}      & \textbf{62.84}      & \smash{\underline{67.73}}  \\ \bottomrule[1.0pt]
	\end{tabular}
	\caption{Results for the cross-domain ASTE task: The best results are highlighted in bold, while the second best are underlined. All other results were sourced from \citet{deng2023bidirectional}}
	\label{tab:table4}
\end{table*}
\subsection{Baseline Models}

In our comparative analysis, we evaluate our proposed method against a diverse array of strong baseline methods, categorized into two main groups:

1) \textbf{Generation-based models}: This category encompasses a variety of generative approaches to ABSA tasks, including GAS, Paraphrase, UIE, DLO, Seq2Path, LEGO-ABSA, MvP, and BGCA. These methods span a range of strategies, from basic seq2seq formulations (GAS) and sentiment tuple transformations (Paraphrase) to pre-training frameworks such as UIE, target data augmentation and evaluation of several prompting templates (DLO), tree path predictions (Seq2Path), task prompt design using T5 sentinel tokens (LEGO-ABSA), multi-view prompting for data augmentation (MvP), and bidirectional generative models for data augmentation (BGCA), offering a comprehensive overview of current approaches in the field.

2) \textbf{Other models}: This group includes span-level models like Span-ASTE, reading comprehension-based models such as BMRC and RoBMRC, alongside discriminative techniques like Extract-Classify. Additionally, TAGS incorporates a sequence labeling aid into the generative model.

\subsection{Evaluation Metrics}

We use the F1 score as the main evaluation metric for assessing the E2TP two-step strategy. The reported F1 scores for all datasets are averaged over 4 different runs. A predicted tuple is considered accurate solely if every element matches precisely with those of the gold tuple.

\subsection{Tasks and Datasets}

The E2TP strategy is assessed across four tasks using datasets from two main domains. In dataset-specific setups, ‘Rest (R)’ denotes the restaurant domain, while ‘Lap (L)’ denotes the laptop domain. The tasks comprise two quadruplet tasks: ASQP, utilizing two datasets from SemEval tasks \citep{pontiki2015semeval,pontiki2016semeval} aligned and completed by \citet{zhang2021aspect}, and ACOS, focusing on implicit aspects and opinions with two datasets provided by \citet{cai2021aspect}. Additionally, there are two triplet tasks: ASTE, with four datasets developed by \citet{xu-etal-2020-position}, and TASD, with two datasets developed by \citet{wan2020target}. Statistical information for these datasets is detailed in Appendix \ref{ap:third}. Cross-domain setups involve transitioning from source to target domains (S$\rightarrow$T) for evaluation. ASTE task datasets are used, employing the BGCA method for domain switching to create six states.

\section{Results and Discussions}

\label{sec:result}
\subsection{Main Results}

The analysis of results emphasizes selecting the best-performing paradigm from three options in each dataset, as well as separately explaining the \(diet\) version.\

Exploring the details in Table \ref{tab:table3} comparing our method with state-of-the-art models, it exhibits a notable average increase in F1 score over MvP, showcasing a substantial average enhancement of \(+1.77\%\). Particularly in the \(diet\) paradigm encompassing all datasets, E2TP achieves an average \(1.35\%\) higher F1 score than MvP, despite having access to less data, notably less than \(35\%\) in quadruplet tasks. Overall, E2TP consistently outperforms MvP across most datasets, with non-diet paradigms consistently superior and the \(diet\) paradigm surpassing MvP in all datasets except one.
E2TP outperforms TAGS, the current state-of-the-art model for the ASTE task, with an average F1 score improvement of \(0.93\%\) even though the fact that, in contrast to E2TP, TAGS is an ASTE task-focused approach. E2TP also surpasses MvP in TASD, ASQP, and ACOS tasks, with average F1 score improvements of \(0.84\%\), \(1.79\%\), and \(0.71\%\) respectively.\
\begin{figure*}[htbp]
	\centering
	\begin{minipage}{0.65\textwidth}
		\centering
		\captionsetup{font=footnotesize}
		\footnotesize
		\begin{tabularx}{0.93\textwidth}{l||*{4}{>{\centering\arraybackslash}c}}
			\toprule[1.0pt] 
			\textbf{Subtasks} & \scriptsize{\textbf{TASD(R15)}} & 
			\multicolumn{1}{l}{\scriptsize{\textbf{ASTE(L14$\rightarrow$R15)}}} & \scriptsize{\textbf{ASQP(R16)}} & \scriptsize{\textbf{ACOS(Lap)}} \\
			\midrule[0.5pt]
			\footnotesize{ATE} & 77.01 & 74.63 & 79.12 & 78.56 \\
			\footnotesize{OTE} & - &  76.17 & 75.38 & 82.79  \\
			\footnotesize{ACD}& 79.05 & - & 81.52 & 56.72 \\ 
			\footnotesize{SPD}& 82.38 &  80.36 & 86.43 & 84.36 \\ 
			\bottomrule[1.0pt]
		\end{tabularx}
		\captionsetup{justification=centering}
		\captionof{table}{Results of the first-step model in aspect term extraction (ATE), opinion term extraction (OTE), aspect category detection (ACD), and sentiment polarity detection (SPD) subtasks.}
		\label{tab:4}
	\end{minipage}
	\begin{minipage}{0.34\textwidth}
		\centering
		\captionsetup{font=footnotesize,justification=centering}
		\includegraphics[width=\textwidth]{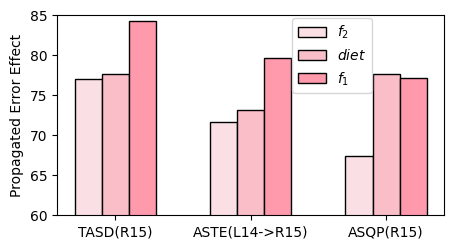}
		\caption{Propagated error effect}
		\label{fig:plt1}
	\end{minipage}
\end{figure*}

\begin{figure}
	\centering
	\includegraphics[width=0.82\columnwidth]{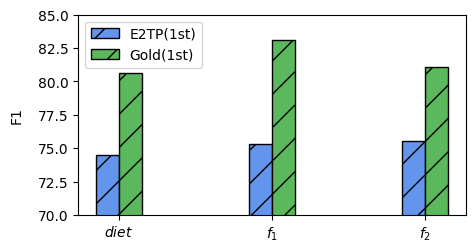}
	\captionsetup{font=footnotesize,justification=centering}
	\caption{Pure analysis of second step model}
	\label{fig:plt2}
\end{figure}
In cross-domain evaluations, as shown in Table \ref{tab:table4}, our method outperforms BGCA across all cases, establishing it as the state-of-the-art model, and our F1 score exceeds BGCA's by \(2.61\%\) on average and by \(1.93\%\) in the \(diet\) paradigm, showcasing E2TP superior performance and adaptability.

\subsection{First Step Model Analysis}
For further analysis of the first step model, we examine two critical cases. The first assesses the performance and F1 scores of first step models on single element extraction tasks (Table \ref{tab:4}). The second, illustrated in Figure \ref{fig:plt1}, delves into the percentage of the propagated error effect, indicating the degree of misguidance caused by first step generated elements in the second step. The percentage reflects the ratio of final wrongly predicted tuples, where only incorrect first step predictions led to their selection (avoidable with the gold elements), to the total of final incorrect tuples. The \(f_1\) paradigm displays a high rate, possibly due to its multiplicity indication technique, switching “$\rightarrow$” and “$\Rightarrow$” symbols, which has a higher potential for error propagation. Moreover, the \(f_2\) and \(diet\) paradigms show a significant rate difference in the quadruplet task, unlike the triplet tasks. This suggests that adapted multi-view prompting could mitigate errors by creating more varied paths.
\subsection{Second Step Model Analysis}
In assessing the second step model, we opt for gold single elements instead of single-step predictions for a more precise evaluation. In Figure \ref{fig:plt2}, a comparison is made where second step outcomes rely on gold outputs for all element types rather than the first step predictions and actual outputs based on the E2TP first step prediction. Despite minor performance differences in real cases, the \(f_1\) paradigm outperforms \(f_2\) and \(diet\) paradigms when gold elements are available, possibly due to the multiplicity indication technique. Although utilizing first step predictions may propagate errors, this technique enhances the pure performance of the second step model. Further analysis of the second step models are available in Appendix \ref{ap:fourth}.
\begin{figure*}[t]
	\centering
	\begin{minipage}{0.54\textwidth}
		\includegraphics[width=\textwidth]{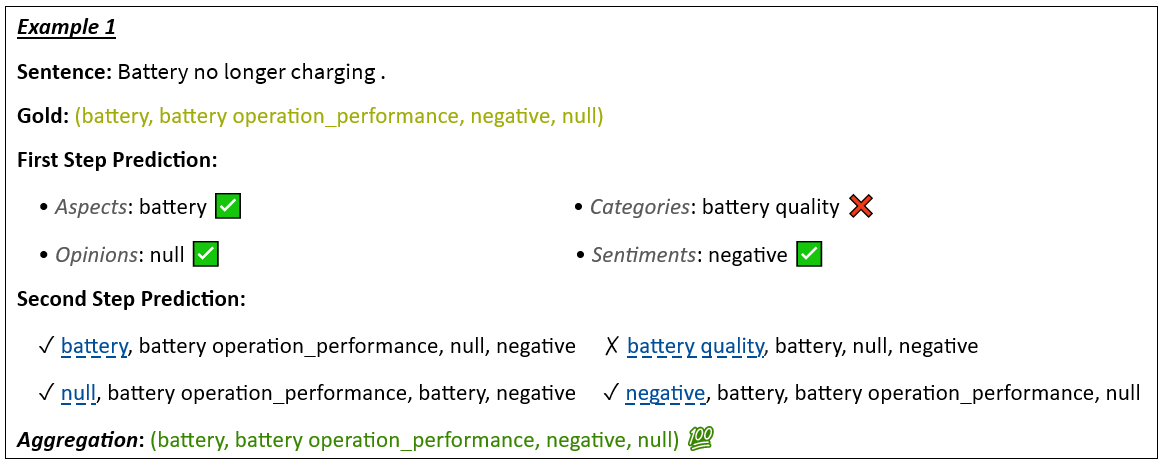}
		\label{fig:figure_label2}
	\end{minipage}
	\begin{minipage}{0.45\textwidth}
		\includegraphics[width=\textwidth]{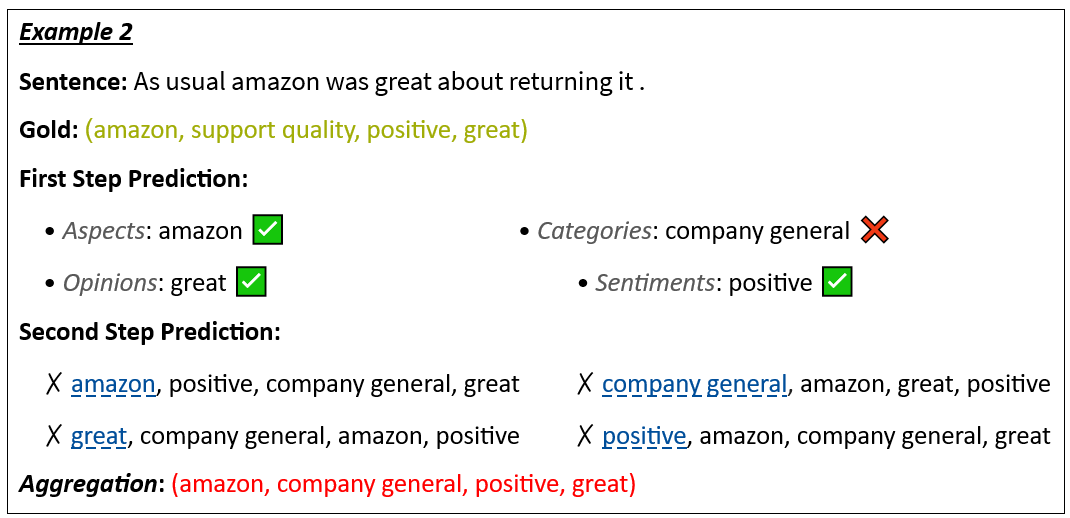}
		\label{fig:figure_label3}
	\end{minipage}
	\captionsetup{font=footnotesize,justification=centering}
	\caption{The case study of E2TP presents the input, model outputs from both steps, and the gold output.}
	\label{fig:case}
\end{figure*}
\subsection{Case Study}
Figure \ref{fig:case} illustrates two examples from the Laptop dataset for the ACOS task within the \(diet\) paradigm. In the first example, despite an initial incorrect prediction of the aspect category and one wrongly-based tuple, E2TP's output in the second step leads to the correct target, showcasing its resilience. However, in the second example, the dataset's complexity, with 121 categories, challenges E2TP to select the most accurate category among multiple possible correct candidates, based on its limited gained knowledge from fine-tuning.

\section{Conclusion}
\label{sec:conclusion}
This study introduces E2TP, a new and straightforward two-step prompting method effectively using data augmentation by employing single elements for tuple prediction. It provides three paradigms, aiming to implement different prompting template styles and create a data-efficient approach capable of making predictions with substantially lower data augmentation rates, while also introducing paradigms with higher data augmentation rates. E2TP outperforms strong baselines in aspect sentiment tuple prediction, showcasing its potential to enhance efficiency and precision in NLP tasks. Our method sets a new benchmark, indicating promising avenues for future research in this field.

\section{Limitations}
\label{sec:limitations}
Despite achieving state-of-the-art performance, our proposed strategy has several limitations that illuminate paths for future research. The two-step procedure and non-diet paradigms, with their increased data augmentation, necessitate more computational resources during training and extend inference time. Though evaluated solely in the ABSA task, the concept of element-to-tuple prompting shows promise for broader application in diverse NLP tasks. Its potential extends to tasks where outputs can be represented as tuples of elements, suggesting avenues for further exploration and development in the field.\

Critical to the refinement of our strategy is the design of effective prompts for each step, aiming to enhance results by meticulously tailoring the prompts to the task at hand. Alongside this, the introduction of a filtering mechanism to minimize errors by preventing the propagation of potentially mispredicted single elements between steps is vital for maintaining the integrity of results. Furthermore, focusing on improving each step and training more expert models, as based on the two-step prompting architecture, the advancement of any step or subtask will improve the overall performance. Additionally, for the \(diet\) version, which currently selects a permutation randomly, the task prompts and the order of target sequence elements are adjusted for all data based on this selection. While the process is random in this project, selecting a permutation that yields better results in a specific dataset would be more appropriate.

\bibliography{custom}
\appendix
\begin{table*}[t]
	\captionsetup{font=footnotesize}
	\centering
	\begin{tabular}{l||*{2}{c}|*{4}{c}|*{2}{c}|*{2}{c}}
		\toprule[1.0pt] 
		\multirow{2}{*}{\textbf{Task}} & \multicolumn{2}{c|}{\textbf{TASD}} & \multicolumn{4}{c|}{\textbf{ASTE}} & \multicolumn{2}{c|}{\textbf{ASQP}} & \multicolumn{2}{c}{\textbf{ACOS}} \\
		& \textbf{R15} & \textbf{R16} & \textbf{L14} & \textbf{R14} & \textbf{R15} & \textbf{R16} & \textbf{R15} & \textbf{R16} & \textbf{Lap} & \textbf{Rest} \\
		\hline
		\\[-3.0ex]
		\footnotesize\textit{\textbf{\underline{ORG\textsuperscript{*}}}} & & & & & & & & & \\
		Train & 1120 & 1708 & 906 & 1266 & 605 & 857 & 834 & 1264 & 2934 & 1530 \\
		Dev & 10 & 29 & 219 & 310 & 148 & 210 & 209 & 316 & 326 & 171 \\
		Test& 582 & 587 & 328 & 492 & 322 & 326 & 537 & 544 & 816 & 583 \\ 
		\hline
		\\[-3.0ex]
		\footnotesize\textit{\textbf{\underline{AUG\textsuperscript{†}}}} & & & & & & & & & \\
		E2TP(\(diet\)) & 4226 & 6422 & 3485 & 5494 & 2423 & 3398 & 4354 & 6543 & 14170 & 7999 \\
		MvP & 5600 & 8540 & 4530 & 6330 & 3025 & 4285 & 12510 & 18960 & 44010 & 22950 \\
		E2TP(\(f_1\)/\(f_2\)) &  8452 & 12844 & 6970 & 10988 &4846 & 6796 & 26124 & 39258 & 85020 & 47994 \\
		\bottomrule[1.0pt]
	\end{tabular}
	\caption{Dataset Statistics for Different Tasks. * \hspace{0.1em}Indicates details of the original data, and † \hspace{0.1em}signifies training data details after running the data augmentation approach of the corresponding method to evaluate the volume of data augmentation of E2TP and MvP strategies. It is important to note that the newly generated data points are solely derived from running the strategies on the original data without accessing any other external unfair data.
	}
	\label{tab:table2}
\end{table*}

\section{Data Statistics}
Table \ref{tab:table2} presents the data statistics of all datasets for the TASD, ASTE, ASQP, and ACOS tasks. To ensure fairness, identical data splits as in previous studies are utilized. E2TP paradigms employ a two-step prompting method, training models concurrently due to their independence at this stage. To analyze E2TP paradigm data usage, we focus solely on the second step, which contains the most data and demands more training time. The \(diet\) paradigm is highly data-efficient, utilizing only 50\% of data in triplet tasks and 16.7\% (\(1/6\)) in quadruplet tasks compared to other paradigms like \(f_1\) or \(f_2\). Even compared to the strong baseline MvP, \(diet\) paradigm requires less data, using less than 35\% in quadruplet tasks. Additionally, the mentioned table depicts a comparison between the data augmentation rates of E2TP paradigms and MvP, as two of the most recent and effective prompting data augmentation strategies.
\label{ap:third}

\section{BGCA Data Augmentation}
\label{sec:bgca}
The BGCA strategy initially enhances data by training a text-to-label model. This model formats inputs like “\(\langle{pos}\rangle \  a\  \langle {opinion}\rangle \  o\)”, where “\(\langle{pos}\rangle\)” denotes sentiment polarity (positive, negative, or neutral), “a” indicates the aspect term, and “o” represents the opinion term , with multiple sentiment tuples separated by spaces. Additionally, it concurrently trains a label-to-text model using the same data but reverses inputs and targets, effectively swapping their positions. After refining both models in the source domain, the text-to-label model is tested on the target domain's test data, and its output is used to augment text data through noisy generated labels. This process transfers knowledge from the target data to the source data without needing access to the target domain's training data.

\section{Detailed Experimental Settings}
\label{sec:imp}
To ensure fairness and align with established practices, our approach integrates the T5-B{\footnotesize ASE} model from the Huggingface transformers library\footnote{\footnotesize{\url{https://github.com/huggingface/transformers}}}, alongside the AdamW optimizer \cite{loshchilov2017decoupled}, across both steps. We use greedy search for decoding in all inference tasks. Regarding hyperparameters, we consistently used a batch size of 16 for all experiments in the training process. All initial step models employed a learning rate of 3e-4. In dataset-specific experiments, such as ASTE(R14\&R15), TASD(R15), and ASQP(R16), the initial step models underwent training for 15 epochs. However, for other datasets, this duration was extended to 20 epochs, and in cross-domain studies, the training duration remained at 15 epochs. All of the second step models were trained for 20 epochs at a learning rate of 1e-4 in both dataset-specific and cross-domain settings across all paradigms and datasets, except for ASQP(R16) dataset in non-diet paradigms, which was trained for 15 epochs. Limited by computational resources, our experiments were exclusively performed on ACOS task datasets using the \(diet\) paradigm. \

Regarding inference hyperparameters for the three paradigms of E2TP, for all experiments involving triplet tasks encompassing both cross-domain and dataset-specific scenarios for the \(diet\) paradigm, the parameters are set to \(m = 1\) and \(d = 0\). However, for the non-diet versions, these parameters change to \(m = 3\) and \(d = 1\). In the case of all quadruplet tasks under the \(diet\) paradigm, the settings are \(m = 2\) and \(d = 1\), while for the non-diet paradigms, the parameters are adjusted to \(m = 11\) and \(d = 2\).
All models trained on a single NVIDIA Tesla T4 16GB GPU.

\begin{table}[t]
	\centering
	\captionsetup{font=footnotesize}
	\begin{tabular}{@{} l||ccc @{}} 
		\toprule[1.0pt] 
		\textbf{Gold}& \scriptsize\textbf{TASD(R15)} & \scriptsize\textbf{ASTE(L14$\rightarrow$R15)} & \scriptsize\textbf{ASQP(R16)} \\
		\midrule
		aspect & 69.25 & 65.36 & 64.57 \\
		opinion & - & 66.76 & 64.68 \\ 
		category & 67.82 & - & 63.30 \\
		sentiment & 66.87 & 63.24 & 63.39 \\
		\midrule
		E2TP & 65.80 & 62.84 & 62.90 \\
		\bottomrule[1.0pt] 
	\end{tabular}
	\captionsetup{font=footnotesize,justification=centering}
	\caption{Improvements in E2TP results through utilizing the  gold elements tailored to the corresponding element type}
	\label{tab:table_label3}
\end{table}

\section{Further Second Step Model Analysis}
\label{ap:fourth}
For a further analysis of the second step models, demonstrating that enhancing any subtask can improve final predictions, Table \ref{tab:table_label3} represents the F1 scores achieved by running the second step model using the gold elements corresponding to the mentioned particular type, and employing normal first step predictions for elements with other types. This illustrates the effectiveness of our decomposition and two-step method, which possesses a proper modularity feature. In this table, TASD(R15), ASTE(L14$\rightarrow$R15), and ASQP(R16) have employed paradigms \(f_1\), \(f_2\), and \(diet\) respectively.

\end{document}